\documentclass[journal]{IEEEtranTIE}
\usepackage{graphicx}
\usepackage{cite}
\usepackage{picinpar}
\usepackage{amsmath}
\usepackage{url}
\usepackage{flushend}
\usepackage[utf8]{inputenc}
\usepackage{colortbl}
\usepackage{soul}
\usepackage{multirow}
\usepackage{pifont}
\usepackage{color}
\usepackage{alltt}
\usepackage{enumerate}
\usepackage{siunitx}
\usepackage{breakurl}
\usepackage{epstopdf}
\usepackage{pbox}

\usepackage{tabu}                     
\usepackage{multirow}                 
\usepackage{multicol}                 
\usepackage{multirow}                
\usepackage{float}                    
\usepackage{makecell}                 
\usepackage{booktabs}                 

\usepackage{graphicx}
\usepackage{cite}
\usepackage{picinpar}
\usepackage{amsmath}
\usepackage{url}
\usepackage{flushend}
\usepackage{colortbl}
\usepackage{soul}
\usepackage{multirow}
\usepackage{pifont}
\usepackage{color}
\usepackage{alltt}
\usepackage{enumerate}
\usepackage{siunitx}
\usepackage{breakurl}
\usepackage{epstopdf}
\usepackage{pbox}
\usepackage[vlined,ruled]{algorithm2e}
\usepackage{caption}
\usepackage{subcaption}
\usepackage{booktabs} 
\usepackage{multirow}  
\usepackage{xcolor}
\usepackage{threeparttable}
\usepackage{graphics}
\usepackage{amssymb}
\usepackage{booktabs}
\usepackage{graphicx}
\usepackage{cite}
\usepackage{picinpar}
\usepackage{amsmath}
\usepackage{url}
\usepackage{flushend}
\usepackage{colortbl}
\usepackage{soul}
\usepackage{multirow}
\usepackage{pifont}
\usepackage{color}
\usepackage{alltt}
\usepackage{enumerate}
\usepackage{siunitx}
\usepackage{breakurl}
\usepackage{epstopdf}
\usepackage{pbox}
\usepackage[vlined,ruled]{algorithm2e}
\usepackage{caption}
\usepackage{subcaption}
\usepackage{booktabs} 
\usepackage{multirow}  
\usepackage{xcolor}
\usepackage{threeparttable}
\usepackage{graphics}
\usepackage{amssymb}
\usepackage{booktabs}
\usepackage[hidelinks,colorlinks,bookmarksopen,bookmarksnumbered,citecolor=green, linkcolor=red, urlcolor=green]{hyperref}

\begin{document}
\title{	GVD-TG: Topological Graph based on Fast Hierarchical GVD Sampling for Robot Exploration }

\author
{
	\vskip 1em
	
	Yanbin Li, Canran Xiao, Shenghai Yuan, Peilai Yu, Ziruo Li, Zhiguo Zhang, Wenzheng Chi* and Wei Zhang*

	\thanks
{

This project was supported by National Natural Science Foundation of China grant \#62273246 awarded to Wenzheng Chi and by the Fund of State KeyLaboratory of IPOC (BUPT) under Grant IPOC2025ZT07. 
   
Yanbin Li, Wei Zhang and Zhiguo Zhang are with School of Electronics Engineering, Beijing University of Posts and Telecommunications, Beijing, 100876, China (yanbinli@bupt.edu.cn; weizhang13@bupt.edu.cn; zhangzhiguo@bupt.edu.cn;).
Canran Xiao is with the School of Cyber Science and Technology, Shenzhen Campus of Sun Yat-sen University, Shenzhen, China (xiaocanran999@gmail.com).
Shenghai Yuan is with the School of Electrical and Electronic Engineering, Nanyang Technological University, 50 Nanyang Avenue, Singapore 639798 (shyuan@ntu.edu.sg).
Peilai Yu is with the Institute for Computer Science, Ludwig Maximilian University of Munich, Munich, Germany (peilai.yu@campus.lmu.de).
Ziruo Li is with Key Lab of Smart Agriculture Systems, China Agricultural University, Beijing 100083, China (liziruo123@cau.edu.cn).
Wenzheng Chi is with Robotics and Microsystems Center, School of Mechanical and Electric Engineering, Soochow University, Suzhou, 215021, Jiangsu, China (wzchi@suda.edu.cn).

    $^{*}$ corresponding author
}

}

\maketitle

\begin{abstract}
Topological maps are more suitable than metric maps for robotic exploration tasks. However, real-time updating of accurate and detail-rich environmental topological maps remains a challenge. This paper presents a topological map updating method based on the Generalized Voronoi Diagram (GVD). First, the newly observed areas are denoised to avoid low-efficiency GVD nodes misleading the topological structure. Subsequently, a multi-granularity hierarchical GVD generation method is designed to control the sampling granularity at both global and local levels. This not only ensures the accuracy of the topological structure but also enhances the ability to capture detail features, reduces the probability of path backtracking, and ensures no overlap between GVDs through the maintenance of a coverage map, thereby improving GVD utilization efficiency. Second, a node clustering method with connectivity constraints and a connectivity method based on a switching mechanism are designed to avoid the generation of unreachable nodes and erroneous nodes caused by obstacle attraction. A special cache structure is used to store all connectivity information, thereby improving exploration efficiency. Finally, to address the issue of frontiers misjudgment caused by obstacles within the scope of GVD units, a frontiers extraction method based on morphological dilation is designed to effectively ensure the reachability of frontiers. On this basis, a lightweight cost function is used to assess and switch to the next viewpoint in real time. This allows the robot to quickly adjust its strategy when signs of path backtracking appear, thereby escaping the predicament and increasing exploration flexibility. And the performance of system for exploration task is verified through comparative tests with SOTA methods. 
The source code will be available upon acceptance at:
https://github.com/littleBurgerrr/Hierarchical\_GVD\_Explora\\tion.git.
\end{abstract}



{}

\definecolor{limegreen}{rgb}{0.2, 0.8, 0.2}
\definecolor{forestgreen}{rgb}{0.13, 0.55, 0.13}
\definecolor{greenhtml}{rgb}{0.0, 0.5, 0.0}



\begin{figure}[t]
\centerline{\includegraphics[width=0.48\textwidth]{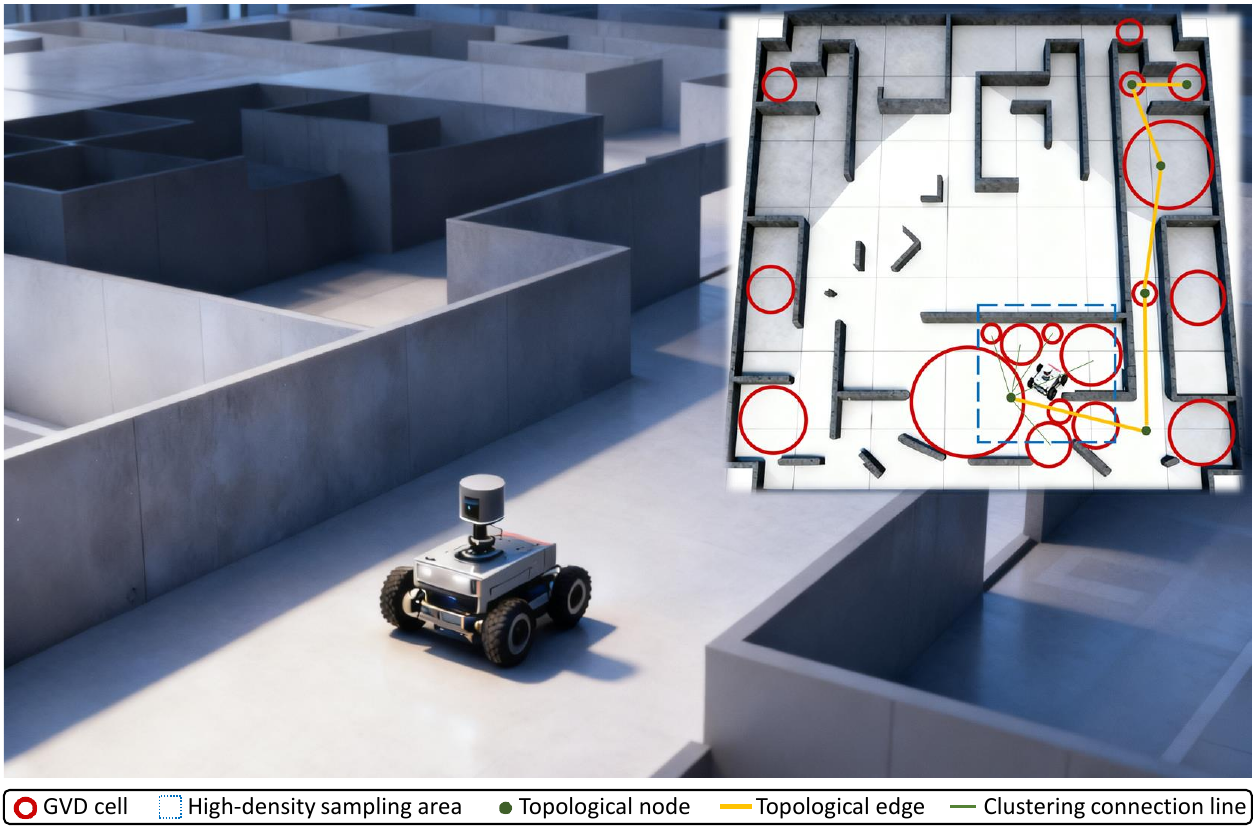}}
\caption{To balance the detail of grid maps and the efficiency of topological maps, we propose an exploration framework based on dynamic hierarchical GVD, which avoids path backtracking and improves exploration efficiency by maintaining accurate topological details in real time.}
\label{ex_show}
\end{figure}

\section{Introduction}





Autonomous exploration is essential for mobile robots to navigate unknown environments and has broad applications, including post-disaster rescue \cite{delmerico2017active, niroui2019deep, balta2017integrated, casper2003human, wu2015pomdp, kohlbrecher2015human}, agricultural monitoring \cite{christiansen2017designing}, and warehousing logistics. In disaster zones, robots must rapidly map the environment with limited communication and no prior maps to aid rescue efforts. In dynamic warehouses, robots need to continuously update maps to adapt to changing layouts and ensure uninterrupted operations. These scenarios demand exploration strategies that are both highly efficient and capable of providing comprehensive environmental coverage within strict time and energy constraints.

Existing mainstream exploration frameworks generally use occupancy grids as the representation of the environment, which depict each cell in terms of the probability of being occupied by obstacles and thus can finely preserve the geometric details of the environment. However, in large-scale or long-term tasks, the number of grids grows linearly, causing the global optimization to be solved in an extremely high-dimensional state space, with the computational complexity increasing exponentially. As a result, the system is forced to degrade to using local greedy strategies, which leads to the problem of path backtracking for robots. Meanwhile, the huge storage and search costs significantly increase the time consumption of replanning. Local obstacles that trigger replanning further exacerbate the risk of backtracking. In contrast, topological maps abstract the environment with a node-edge structure, which naturally fits the hierarchical architecture. That is, the global planner first quickly screens feasible regions at the topological level, and then the local planner refines the continuous trajectory. It is easier to identify and skip narrow branches in advance, thereby effectively suppressing backtracking. In addition, since the number of topological nodes is much smaller than that of grid cells, topological maps have higher real-time performance in large-scale or continuous exploration tasks and can quickly get out of trouble through lightweight node reconnection and edge weight updates.

The key challenge is how to maintain an accurate and detail-rich topological map in the long term. Despite the significant advantages of topological maps in suppressing backtracking and improving global efficiency, there are still problems in directly using them for autonomous exploration. Since topological nodes only retain the connectivity skeleton and lack environmental details, robots are prone to missing narrow passages or small objects, causing perceptual blind spots and thus limiting coverage. Secondly, the reliability of topological construction and maintenance is insufficient. A single misdetection may produce a broken edge, and existing methods lack an efficient error correction mechanism, causing the topological map to drift and even fail over time.

To address this issue, this paper proposes a real-time topological update and exploration framework based on the Generalized Voronoi Diagram (GVD). First, a hierarchical, non-overlapping adaptive GVD generation strategy is proposed to accurately extract environmental feature. By dividing the denoised incremental map into global and local regions and applying different sampling frequencies to them, the accuracy of topological nodes is ensured while enriching the capture of environmental details in the local region around the robot and suppressing the generation of redundant nodes to improve efficiency. Second, a mean shift clustering method with connectivity constraints is designed and accelerated by KD-tree to ensure that the generated topological nodes are always located in free space and avoid cross-obstacle attraction. The connectivity of topological nodes is ensured by a dual-layer switching connection mechanism, ensuring the completeness and efficiency of regional connectivity. Finally, by integrating the topological length, turning cost, and end deviation into a lightweight cost function, the real-time optimal selection of boundary viewpoints is realized. In addition, robots can dynamically reselect the next target according to the latest topological update during the movement, laying the foundation for timely escape and continuous efficient exploration.
The primary contributions are:

\begin{enumerate}[1)]
	\item Efficient environmental modeling integrating incremental noise reduction and hierarchical GVD sampling:
    An innovative environmental feature extraction process is proposed. Firstly, local trusted neighborhood voting denoising is designed for the incremental map to eliminate perceived noise; Furthermore, a hierarchical coverage-aware GVD sampling strategy is adopted to achieve adaptive sampling of local refinement and global sparsity. This design ensures the accuracy of the topological structure and the efficiency of the calculation at the source.
    
	\item Topology diagram construction method with connectivity constraints:
    A Mean Shift clustering algorithm combined with connectivity constraints and a double-layer switching connection mechanism are designed to ensure that all topological nodes are located in the reachability space, and the GVDs or bidirectional A* algorithm can be adaptively selected to guarantee the global connectivity and reliability of the topological diagram.


    \item Morphological frontier detection and real-time viewpoint switching:
    Innovatively utilize morphological operations to quantify the connectivity between unknown regions and free space, effectively filtering out pseudo-boundary points. It is supplemented by a lightweight cost function, enabling the robot to dynamically select the optimal viewpoint based on the real-time updated topological diagram, significantly reducing ineffective exploration and path backtracking.
	
\end{enumerate}

\section{Related work}

\subsection{Metric Map-based Exploration}

Liu $\textit{et al.}$ \cite{9893536} biased sampling within the free voxels of the occupancy grid via heuristically weighted random points to accelerate indoor robot exploration.
Batinović $\textit{et al.}$ \cite{9387089} identified frontiers in a multi-resolution 3-D occupancy octree, refining voxel clusters progressively to balance accuracy and computation for aerial exploration.
Zhou $\textit{et al.}$ \cite{9324988} presented FUEL, it maintained an incremental frontier information structure (FIS) using a voxel grid map and employed hierarchical planning to achieve efficient exploration.
Zhang $\textit{et al.}$ \cite{10423847} clustered frontier cells from a 3-D occupancy grid into macro-frontiers and selected the next target via an information-gain-over-cost metric to accelerate UAV mapping.
Zhao $\textit{et al.}$ \cite{10155653} presented an autonomous exploration method for UAVs with limited FOV sensors. They generated and evaluated spiral trajectories in a rolling 2-D occupancy grid based on frontier visibility and path cost, selecting the best trajectory to enhance exploration efficiency.
The aforementioned exploration methods based on gridded or voxelized metric maps face substantial burdens in terms of storage, computation, and memory in high-dimensional or large-scale environments, with the magnitude of these issues increasing exponentially as resolution improves. To maintain real-time performance, algorithms often employ coarse resolutions, which in turn increase localization errors and collision risks. Moreover, the high computational complexity, low update frequency, and lack of global structural information in metric maps lead to low path planning efficiency and a propensity for path backtracking, particularly in complex environments.

\subsection{Heuristic Methods based on Topological Information}

Some methods leverage the topological information of the environment to heuristically guide the robot's path planning, enhancing exploration efficiency by extracting and utilizing the structured information of the environment.
Chen $\textit{et al.}$ \cite{10713282} extracted a topological skeleton from a 2-D occupancy grid and used it as a homotopy-heuristic guide within an RRT* variant to focus sampling only where refinement is necessary.
Liu $\textit{et al.}$ \cite{9893536} biased sampling within the free voxels of the occupancy grid via heuristically weighted random points to accelerate indoor robot exploration.
Cao $\textit{et al.}$ \cite{cao2021tare} presented a hierarchical exploration framework that employs a topological approach by partitioning the 3D environment into subspaces for global planning and uses local RRT* within selected subspaces to efficiently explore complex 3D environments.
Soroya $\textit{et al.}$ \cite{8968613} proposed an active SLAM method that constructs a connectivity graph from a probabilistic occupancy map. The graph captures the topological structure of the environment, and its edge weights guide the robot to maximize loop-closure probability during exploration.
Yang $\textit{et al.}$ \cite{10314737} proposed a method for multi-agent exploration using topological graphs derived from occupancy grids. Graph neural networks are trained to predict next-best nodes for agents, refining the joint map during decentralized exploration.
Zhang $\textit{et al.}$ \cite{9834084} developed a multi-robot exploration technique that builds local topological maps within each robot’s occupancy grid and merges them into a global topology during intermittent communication, facilitating exploration under bandwidth limitations.
Huang $\textit{et al.}$ \cite{10015689} presented a fast large-scale ground-robot exploration method. It detects frontiers in a 2-D occupancy grid, clusters them into a topological graph, and solves a TSP-like optimization to select the next best viewpoint.
However, these heuristic methods based on topological information heavily rely on the accuracy of the heuristic module. Moreover, these methods often require prior knowledge to extract and utilize topological information, such as predefined environmental features like rooms and corridors. These features are difficult to obtain in completely unknown environments, thereby limiting their applicability. Meanwhile, they typically assume that the environment has structural characteristics, which restricts their application in unstructured environments.

\subsection{Topological Map-based Exploration}

Topological maps address the aforementioned issues by extracting the key structures of the environment and retaining only the essential nodes and connections. This approach significantly reduces the consumption of storage and computational resources and enables the rapid generation of navigation paths, making it suitable for real-time applications. Moreover, topological maps provide global layout information of the environment, assisting robots in understanding environmental connectivity and achieving efficient long-range planning, thereby avoiding path backtracking problems.
Zhao $\textit{et al.}$ \cite{9861387} extracted a skeleton graph from a global occupancy grid and applied graph partitioning plus shortest-path queries to build a consistent topological map for multi-robot coordination.
Saroya $\textit{et al.}$ \cite{9387607} developed a topology-informed Growing Neural Gas algorithm to generate sparse navigation roadmaps in uncertain environments. The method uses persistent homology to identify key topological features, guiding the roadmap generation to focus on challenging regions.
Wang $\textit{et al.}$ \cite{8645716} built an incremental roadmap directly inside a 2-D occupancy grid, inserting nodes and edges as new free space was discovered and selecting the next best view via graph search to guide autonomous exploration.
However, topological maps primarily focus on the connectivity of the environment and key nodes, and they do not provide precise descriptions of the specific geometric shapes and the exact locations of obstacles. This limitation makes it difficult to explore the details of the environment, thereby reducing the completeness of the exploration task. Moreover, when generating coverage paths, methods based on topological maps do not sufficiently consider the safe distance between the path and obstacles, which may lead to collision risks in practical applications.

\subsection{GVD}

Relying solely on the limited information provided by topological maps, robots struggle to accurately avoid obstacles and determine their next moves. In contrast, the Generalized Voronoi Diagram (GVD) enriches the environmental information available to robots by extracting key geometric structures from the environment, thereby facilitating a better understanding of the layout and aiding in decision-making. The inherent properties of the GVD ensure a safe distance between the robot and obstacles, as the medial axis generated by the GVD naturally provides a collision-avoidance path for the robot.
Chi $\textit{et al.}$ \cite{9336621} reused a GVD-based feature tree extracted from the occupancy grid to provide long-range homotopy classes, guiding an RRT* variant to rapidly escape trapped spaces while retaining completeness.
Chi $\textit{et al.}$ \cite{9430686} built a lightweight GVD skeleton on the 2-D occupancy map and used the resulting feature matrix to bias RRT* sampling toward wide passages, cutting planning time in cluttered mobile-robot scenarios.
Vachhani $\textit{et al.}$ \cite{5605253} computed the GVD directly from range data via a hardware-efficient control-law implementation, producing a sparse medial-axis roadmap for real-time mobile robot navigation without storing the full metric grid.
Chen $\textit{et al.}$ \cite{10771660} extracted a fast GVD from the occupancy grid and fused its medial-axis segments with frontier clusters to create a multi-strategy target assignment, enabling efficient autonomous exploration while leveraging the GVD’s natural connectivity prior.
The existing exploration methods based on GVD mainly have several key drawbacks: Firstly, they adopt a uniform sampling strategy, which makes it difficult to balance global efficiency and local details, resulting in omissions at key structures such as narrow channels. Secondly, the lack of an effective redundant control mechanism will result in a large number of repetitive and ineffective GVD nodes in the explored areas, wasting computing resources. Thirdly, the generated topological nodes are vulnerable to interference from obstacles, which may occur in inaccessible areas or cause incorrect cross-obstacle connections, seriously affecting the reliability of the map. Fourth, the determination of the frontier is rough, often misjudging unknown areas that are inaccessible due to obstacles as targets, resulting in low exploration efficiency.

In response to these deficiencies, this paper proposes a systematic improvement plan: by means of hierarchical adaptive sampling and overlay map maintenance, the balance problem between sampling density and redundancy is solved; The clustering method using connectivity constraints ensures the reachability and accuracy of topological nodes. A frontier detection mechanism based on morphology was designed to effectively screen out truly reachable exploration targets, thereby achieving efficient and reliable autonomous exploration.






\section{Preliminary}

This section aims to provide a theoretical basis for the geometric safety and topological completeness of the exploration method based on GVD.

In the continuous decision-making process of autonomous exploration, the safety of path planning depends on whether the maximum passable margin can be maintained continuously.

GVD is an ideal structure that balances topological integrity and collision risk control. As shown in (\ref{GVD_1}), the GVD sampling point set $V$ is defined as the set of all points that maintain equal distance to at least two obstacles. Among them, $o_i$ and $o_j$ are two different obstacle coordinates in the obstacle set $O$, and $v$ is the sampling point coordinate of a certain GVD cell, $d$ denotes the minimum Euclidean distance.
\begin{equation}
\label{GVD_1}
V=\left\{v\in Z^{2} \mid \exists o_i,o_j\in O,i\neq j,d(v,o_i)=d(v,o_j) \right\}
\end{equation}
This ensures that every GVD sampling points lies on the medial axis of at least two obstacles, thereby naturally avoiding proximity to any single obstacle boundary. 

To simplify the calculation process, we define the shape of GVD cell as a circle with $v$ as the center and $r$ as the radius. Furthermore, the maximum inscribed circle radius $r(v)$ corresponding to each GVD sampling point $v$ is given by (\ref{GVD_r}), which quantifies the maximum safe area available at that point. Since $r(v)$ is explicitly maximized, when a robot traverses along the $V$, its body contour remains entirely within a free disk of radius $r(v)$, guaranteeing a minimum safety distance of at least $r(v)$ in all directions. This property not only keeps the path away from obstacles but also provides a quantifiable safety buffer to accommodate sensor noise, localization errors, and dynamic obstacles. 
It can thus be concluded that, from both geometric and topological perspectives, the GVD demonstrates clear applicability and superiority in robotic exploration tasks. It offers the robot quantifiable and extensible safe passages.
\begin{equation}
\label{GVD_r}
r(v)= \max \left\{ r \mid B(v,r) \cap o  = \varnothing , \forall o \in O \right\}
\end{equation}
Here, $B(v,r)$is a circle with $v$ as the center and $r$ as the radius.

\section{Methodology}

\begin{figure*}[t]
\centerline{\includegraphics[width=1\textwidth]{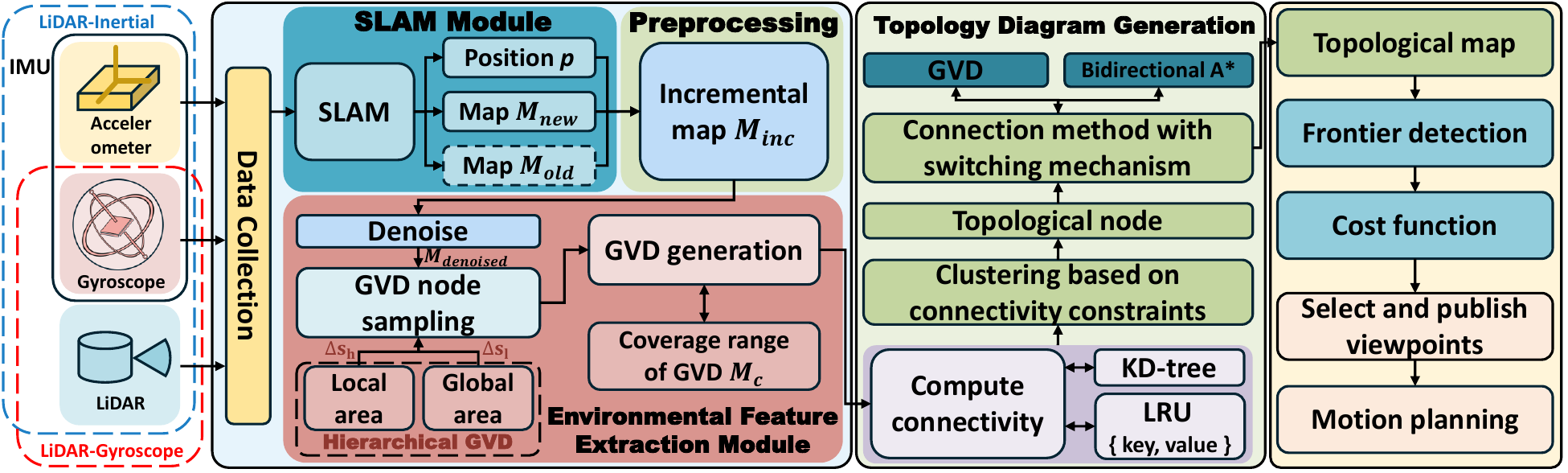}}
\caption{Pipeline.}
\label{pipeline}
\end{figure*}

In this section, we will introduce the proposed methods in detail. Fig. \ref{pipeline} illustrates the pipeline of our system. After receiving sensor data, the robot build environmental map and obtain position using SLAM (Gmapping\cite{grisetti2007improved}). At the next time frame, the system computes the map variation between consecutive frames to generate incremental map $M_{inc}$, which is then denoised using a trusted neighborhood voting based approach to eliminate noise-induced erroneous GVD sampling. And then, a hierarchical GVD sampling strategy is applied to adaptively regulate sampling density, while the GVD coverage matrix $M_c$ is utilized to prevent repeated sampling in overlapping areas, thereby yielding the GVD.
Subsequently, all GVD nodes are clustered via a improved mean shift clustering algorithm based on connectivity constraints to derive topological nodes of the environment. Both GVD nodes and topological nodes are updated frame by frame. After that, a switching mechanism-based topological connection method is employed to generate a complete topological graph. Following topological graph construction, frontier are extracted from all GVD and topological nodes using a morphology-aware detection approach. These frontier are then evaluated by a specially designed cost function to select the next viewpoints.

\subsection{Environmental Feature Extraction Module based on GVD}

\subsubsection{Local Trusted Neighborhood-based Denoising}

\begin{figure*}[htb]
\centerline{\includegraphics[width=1\textwidth]{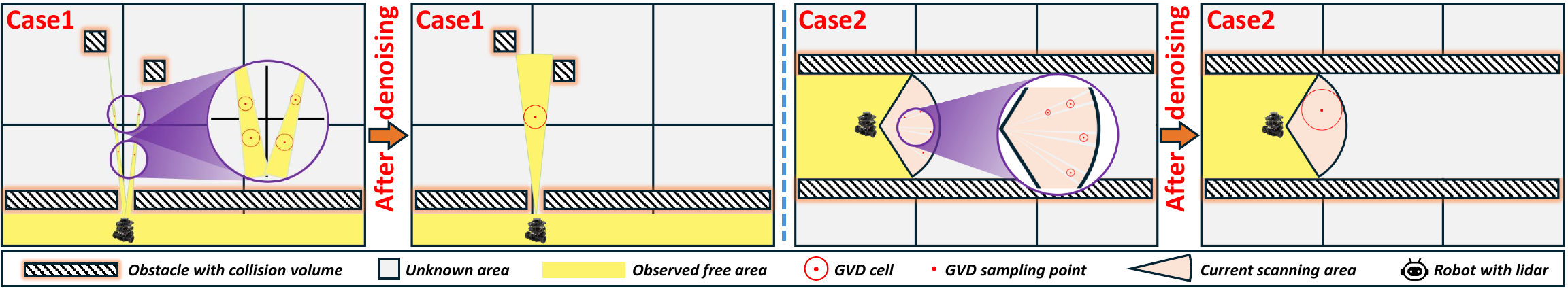}}
\caption{In case 1, LiDAR beams passing through narrow structures create isolated free regions (yellow) separated by large unknown areas (gray). In case 2, limited scanning frequency causes gaps between consecutive frames to be misclassified as unknown space. Both cases produce undersized GVD cells that reduce feature extraction efficiency. With map denoising, fewer GVD nodes are needed to capture environmental topology, significantly improving task performance.}
\label{denoise}
\end{figure*}

Map noise arises from the combined effects of LiDAR frequency and environmental structures. As shown in cases 1 and 2 of Fig. \ref{denoise}, GVD nodes erroneously generated by such noise typically exhibit small coverage areas and information scarcity, which can degrade both the real-time performance of exploration and the accuracy of the topological structure. Therefore, it is necessary to rapidly eliminate these sparse noise points.

To minimize computational time, denoising is applied only to the incremental portion of the map. The specific procedure is outlined in Algorithm \ref{al:denoise}. The difference between the map of the previous time frame $M_{old}$ and the current time frame $M_{new}$ is computed to obtain an incremental map $M_{inc}$ that contains only newly observed data. Subsequently, the minimum bounding rectangle $(x,y,w,h)$ of the newly added region is calculated, and a cropped local map $M_{crop}$ is extracted accordingly. For each incremental grid cell $p$ $\in$ $M_{crop}$ that remains marked as unknown, the number $n$ of grids observed as free within the $5 \times 5$ square neighborhood $N(p)$ centered at $p$ is counted. If $n > 3$, the state of $p$ is updated to free according to a trusted neighborhood voting strategy; otherwise, its original occupancy state is retained. The final output is a denoised map $M_{denoised}$ that applies noise reduction only to the incremental region.

As illustrated in Fig. \ref{denoise}, this approach fully preserves the structure of the prior map while avoiding redundant computation and error propagation caused by noise, enabling the exploration task to extract environmental structures more accurately with fewer GVD nodes.

\begin{algorithm}[t]

\LinesNumbered
\caption{Incremental Map Denoising}
\label{al:denoise}

\KwIn{New map $M_{\mathrm{new}} \subseteq \mathbb{Z}^2=\{p=(x,y)|x,y\in\mathbb{Z}\}$, old map $M_{\mathrm{old}} \subseteq \mathbb{Z}^2=\{p=(x,y)|x,y\in\mathbb{Z}\}$}
\KwOut{Denoised map $M_{\mathrm{denoised}}$}

    Incremental map: $M_{\mathrm{inc}} = M_{\mathrm{new}} - M_{\mathrm{old}}$

    Bounding box: $(x,y,w,h) \gets \textbf{MinimumBoundingRectangle}(M_{\mathrm{inc}})$ 

    Cropped map: $M_{\mathrm{crop}} = \{ p=(u,v) \in M_{\mathrm{new}} | x \leq u < x+w, y \leq v < y+h \}$ 

    \For{$p = (x, y) \in M_{\mathrm{inc}}$}
    {
        \If{$M_{\mathrm{crop}}(p) = \mathrm{unknown}$}
        {
            $\mathcal{N}(p) \gets \{ q \mid q \in \mathbb{Z}^2, \|q- p\|_\infty \leq 2 \}$ 
            
            $n \gets |\{ q \mid q \in \mathcal{N}(p), M_{\mathrm{crop}}(q) \neq \mathrm{unknown} \}|$ 
            
            \If{$n > 3$} 
            {
                $M_{\mathrm{denoised}}(p) \gets \mathrm{free}$
            }
        }
    }
    \Return $M_{\mathrm{denoised}}$
\end{algorithm}

\subsubsection{hierarchical GVD Sampling with Coverage-Aware Prioritization}

\begin{figure}[t]
\centerline{\includegraphics[width=0.5\textwidth]{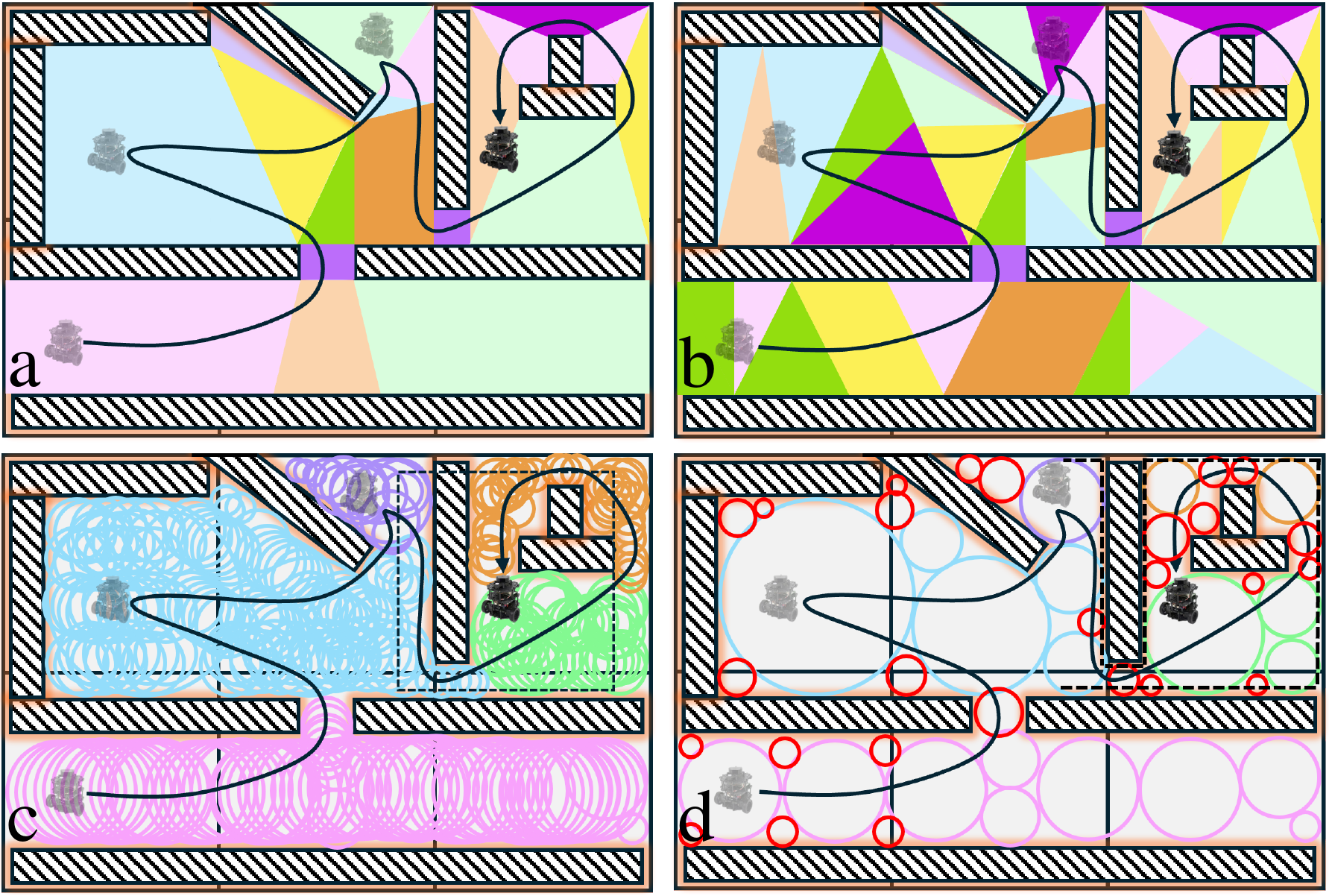}}
\caption{ (a) and (b) illustrate how the sampling density of conventional Voronoi diagrams affects the partitioning of functional regions, while (c) is the regional division effect under high-density GVD sampling. In contrast, (d) shows that our hierarchical sampling strategy achieves broader area coverage with significantly fewer GVD nodes, thereby improving topological accuracy while minimizing computational time.}
\label{hierarchical_gvd}
\end{figure}

The sampling density of the GVD directly determines both the accuracy of the topological graph and the computational efficiency of the entire system. As shown in Fig. \ref{hierarchical_gvd}(a) and Fig. \ref{hierarchical_gvd}(b), the sampling density critically influences how well the GVD captures key structural features such as narrow passages, corners, and bifurcations. Existing GVD-based methods typically use uniform sampling strategies, which face a fundamental trade-off: as illustrated in Fig. \ref{hierarchical_gvd}(c), while high-density sampling preserves environmental details, it introduces significant computational overhead and node redundancy; conversely, low-density sampling tends to omit critical narrow passages, thereby reducing exploration coverage. This limitation becomes particularly pronounced in long-duration exploration tasks, where inefficient sampling can rapidly deplete computational resources.

To address this challenge, we propose hierarchical GVD sampling with coverage-aware prioritization, a method that integrates three key innovations to enable intelligent allocation of sampling resources. First, we design a coverage-aware sampling exclusion strategy. By maintaining a global GVD coverage matrix $M_c$, the system dynamically tracks areas already covered by existing GVD cells. As outlined in Algorithm \ref{al:gvd}, the process begins by taking the denoised map $M_{denoised}$ as input and extracting the set of all free grid cells $p_f$, whose traversal order is randomly shuffled to produce an unordered set $p_s$. Each candidate grid cell $(x, y)$ is then checked against $M_c$ to determine whether it has already been recorded. A GVD node is generated only if it lies in an uncovered area, ensuring that each new node provides unique spatial information and maximizes the information gain per sample.

Second, we establish a dual-grained spatial prioritization mechanism. Based on the robot’s real-time position and following (\ref{Mhigh}), the environment is dynamically divided into a local high-frequency sampling region $M_{high}$ and a global low-frequency sampling region $M_{low}$ (all areas outside $M_{high}$). Within $M_{high}$, a fine step size $\Delta s_{h}$ is used to capture detailed structural features necessary for decision-making, whereas in $M_{low}$, a coarse step size $\Delta s_{l}$ is applied to reduce computational load while maintaining topological consistency.
\begin{equation}
\label{Mhigh}
M_{high} = \left\{ (x,y) \middle| 
\begin{aligned}
& x \in \left[x_r - \frac{L}{2\rho},\ x_r + \frac{L}{2\rho}\right], \\
& y \in \left[y_r - \frac{L}{2\rho},\ y_r + \frac{L}{2\rho}\right]
\end{aligned}
\right\}
\end{equation}
Here, $L$ denotes the search range size and $\rho$ represents the map resolution. Their values are adjusted such that the local region corresponds to a $5 \times 5$ grid area. The robot's position is denoted as $p_r = (x_r, y_r)$.

Third, we introduce an efficiency-oriented GVD generation criterion. To determine the effective range of each GVD cell, the search radius $r$ is incrementally expanded from initial value $r_{min}$ with a step size $\Delta r$. The Manhattan distance $d_m$ defined in (\ref{Manhattan}) is used to rapidly estimate obstacle locations until an obstacle is first detected within scanning region $A$. 
\begin{equation}
\label{Manhattan}
d_m(p_s,p_o) = |x - x_o| + |y - y_o|
\end{equation}
Among them, $p_o=(x_o,y_o)$ is the coordinate of the obstacle.

At this point, the Euclidean distance $d(p_s, p_o)$ shown in (\ref{Euclidean}) to the nearest obstacle is recorded, and all equidistant obstacles $O$ are collected as shown in (\ref{O_near}). 
\begin{equation}
\label{Euclidean}
d(p_s,p_o) = \sqrt{(x - x_o)^2 + (y - y_o)^2}
\end{equation}
\begin{equation}
\label{O_near}
O = \left\{ o_i \mid d_M(p, o_i) < d_{min} + \delta \right\}
\end{equation}
If $|O| > 1$ and the Euclidean distance $d_o$ between obstacles exceeds a predefined threshold, the multi-obstacle bisection criterion is applied to compute the GVD node coordinates $(x_g, y_g)$ and its maximum inscribed circle radius $r_g$, ensuring that the node lies in the safe medial region between multiple obstacles. Nodes $(x_g, y_g)$ satisfying $r_g > r_{min}$ and their corresponding radii $r_g$ are recorded into $\mathcal{V}_{GVD}$ and $\mathcal{R}_{GVD}$, respectively, to facilitate rapid retrieval during subsequent topological node generation. Finally, following (\ref{M_c}), coverage regions are marked in $M_c$ according to radius priority, and the updated GVD is published in real time, providing a zero-redundancy topological partition with practical navigational value.
\begin{equation}
\label{M_c}
M_c = \left\{ (x,y) \mid \sqrt{(x - x_c)^2 + (y - y_c)^2} \leq \lceil r \rceil + 1 \right\}
\end{equation}

The refined sampling mechanism described above enables high-frequency scanning around the robot to capture environmental details, while low-frequency scanning is applied in other areas to reduce computational load. Combined with the incremental GVD coverage matrix and its labeled access mechanism, the system ensures that each node contributes maximally to exploration gains under limited sampling budgets. As shown in Fig. \ref{hierarchical_gvd}c and Fig. \ref{hierarchical_gvd}d, this strategy achieves a dual optimization of both accuracy and efficiency, significantly reducing redundant nodes and path backtracking. During incremental robotic exploration, the GVD adaptively supplements nodes as new areas are discovered, thereby continuously enriching and correcting the environmental topology.

\begin{algorithm}[t]
\LinesNumbered
\caption{GVD}
\label{al:ggd}

\KwIn{Denoised map $M_{\mathrm{denoised}} \subseteq \mathbb{Z}^2$}
\KwOut{GVD node set $\mathcal{V}_{GVD}$, node radius set $\mathcal{R}_{GVD}$}

    Initialize coverage map $M_c$ \\
    
    \While{system is running}
    {
        Find free grids: $p_f = \{ (x,y) \in \mathbb{Z}^2 | M_{\mathrm{denoised}}(x,y)=\text{free} \}$  \\
        
        Shuffle sampling order: $p_s \gets \textbf{Shuffle}(p_f)$  \\
    
        Get robot position: $p_r \gets \textbf{GetPosition}()$  \\
        
        Local map around robot with high sampling density: $M_{\text{high}} \gets \textbf{LocalRange}(p_r)$ \\
    
        \For{$(x,y)$ in $p_s$ with step $\Delta s_h$} 
        {
            \If{$(x,y)$ in $M_c$}
            {
                \textbf{continue}
            }
            \If{($(x,y)$ not in $M_{\text{high}}$) and (not global sampling step $\Delta s_l$)}
            {
                \textbf{continue}
            }
    
            Search radius $r \gets r_{\text{min}}$ \\
            
            \While{no obstacle found}
            {
                Build search matrix $A$ centered at $(x,y)$ with radius $r$  \\
                
                \If{obstacle exists in $A$}
                {
                    \textbf{break}
                }
                
                $r = r + \Delta r$ \\
            }
    
            Compute distance to nearest obstacles $d$ \\
            
            Select nearest obstacle set $O$  \\
            
            \If{$|O| > 1$}
            {
                \For{each obstacle $o_j \in O$}
                {
                    \If{$d_o > \text{threshold}$}
                    {
                        Compute GVD node ($x_g$, $y_g$)  \\
                        
                        Compute radius $r_g$ \\
                        
                        \If{$r_g > r_{\text{min}}$}
                        {
                            $\mathcal{V}_{GVD} \gets \{(x_g, y_g)\}$ \\
                            $\mathcal{R}_{GVD} \gets \{r_g\}$ \\
                            
                        }
                        \textbf{break}
                    }
                }
            }
            Update $M_c$ with node with the largest radius
        }
    }

    \textbf{return} $\mathcal{V}_{GVD}$ and $\mathcal{R}_{GVD}$
\end{algorithm}

\subsection{Topology Diagram Generation and Maintenance Module}

\subsubsection{Connectivity-Constrained Mean Shift Clustering with KD-Tree Acceleration}

\begin{figure}[t]
\centerline{\includegraphics[width=0.5\textwidth]{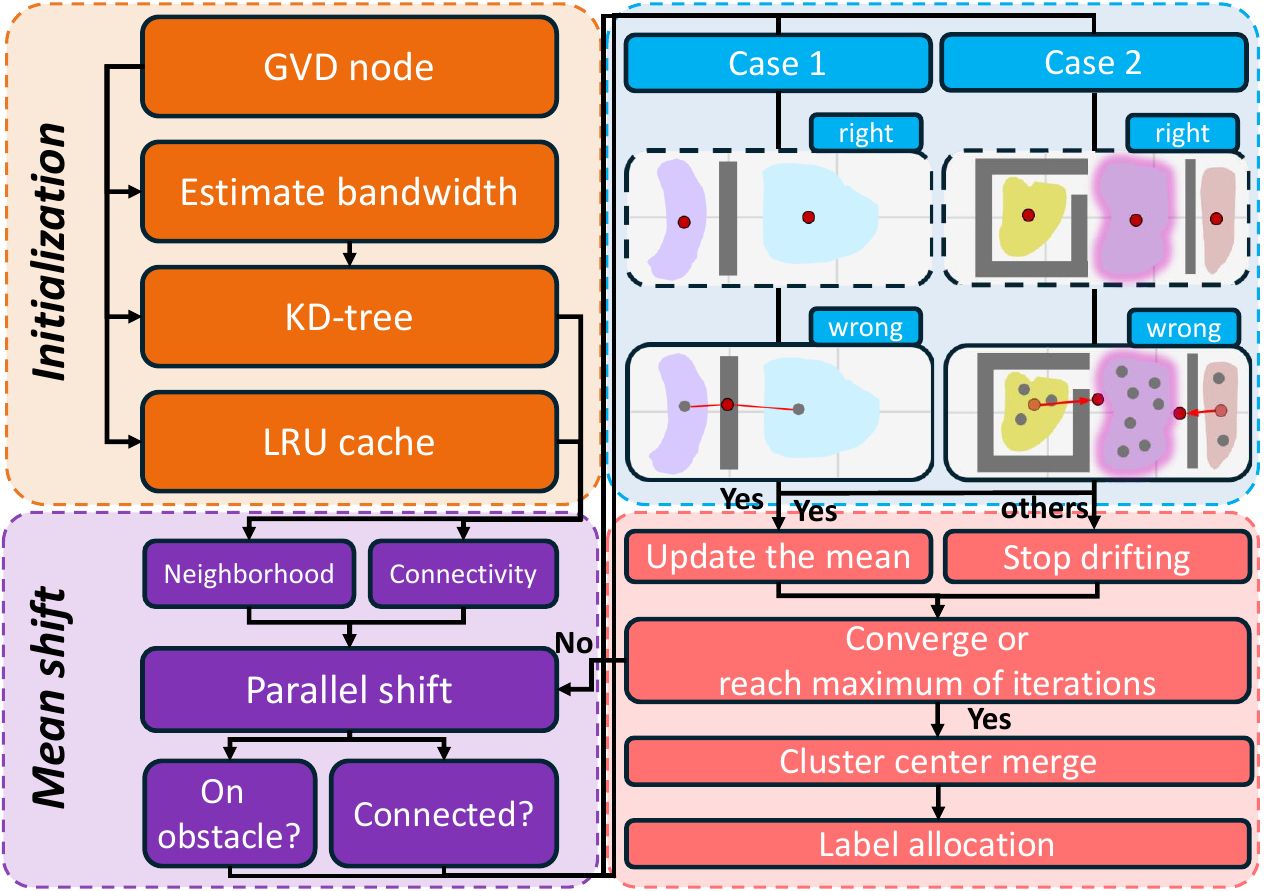}}
\caption{Mean shift based on connectivity constraints. In case 1 and case 2, the red dots are topological nodes and the gray dots are GVD nodes. The correct situation should be that the GVD nodes in each color area are clustered into one topological node. Error situation in case1 using traditional mean shift: The topological node generated by the clustering of two GVD nodes is generated on an obstacle. Error situation in case2 using traditional mean shift: The original correct topological nodes in the GVD areas of yellow and skin color were attracted to the pink area by crossing boundaries.}
\label{mean_shift}
\end{figure}

The traditional mean shift clustering algorithm exhibits three significant limitations in the context of topology construction, as illustrated in case 1 and case 2 of Fig. \ref{mean_shift}. First, cluster centers may fall into obstacle regions, rendering the generated topological nodes unreachable in the physical environment. Second, GVD nodes separated by obstacles may be incorrectly merged due to density attraction, leading to loss of local topological nodes and coverage gaps. Finally, as the environmental scale increases, the number of GVD nodes grows rapidly. If a linear traversal method is still used for neighborhood search, the computational complexity increases sharply, significantly degrading real-time performance and constraining online deployment capability.

To address the aforementioned issues, we propose an improved mean shift framework based on connectivity constraints, which systematically integrates map reachability information into the entire iterative process of mean shift. Furthermore, for the first time in the mean shift algorithm, we introduce KD-tree as the core spatial indexing structure to alleviate the computational burden caused by large-scale GVD nodes. The specific procedure is as follows: taking the set of GVD sample points as input, recursive spatial partitioning is performed on the two-dimensional plane. The root node is split based on the median of the x-coordinates, the next level switches to the y-coordinates, and the subsequent level returns to the x-coordinates, alternating sequentially until the number of points contained in a leaf node falls below a preset threshold. This strategy ensures that the height of the KD-tree remains at $O(\log N)$, reducing the average complexity of any radius-based neighborhood query from $O(N)$ for linear traversal to $O(\log N)$, thereby ensuring real-time responsiveness for subsequent iterations.

To prevent the generation of unreachable cluster centers and suppress cross-obstacle attraction, we deeply integrates connectivity constraints into the mean shift iterative process. The algorithm precomputes connectivity between any two nodes by performing collision-free line segment detection on the occupancy grid map along the straight line connecting them. If the entire segment lies within free space, the two nodes are considered connected; otherwise, they are deemed disconnected. All detection results are stored as key-value pairs in an LRU cache, establishing a multi-resolution, shareable connectivity caching system that avoids redundant computations. 
Subsequently, the bandwidth parameter $r_m$ is adaptively determined according to (\ref{bandwith}): the Euclidean distances between all point pairs in the input dataset are computed, and a threshold from the distance distribution is selected based on a specified quantile to serve as the final bandwidth parameter. The distance value corresponding to this quantile is used as the radius of the neighborhood in the mean shift clustering process.
\begin{equation}
\label{bandwith}
r_m = \text{Quantile}_{q}\left( \left\{ ||\mathcal{V}_i - \mathcal{V}_j|| \mid 1 \leq i < j \leq n \right\}\right)
\end{equation}
where $q$ denotes the quantile parameter and is set to $q = 0.3$, $n$ represents the total number of GVD nodes, and $\mathcal{V}_i$ and $\mathcal{V}_j$ refer to the $i$-th and $j$-th nodes in $\mathcal{V}_{GVD}$, respectively.

Subsequently, all sampling points are used as initial seeds and undergo iterative mean shift. During the iteration, the algorithm first employs the KD-tree to rapidly retrieve the set of neighborhood points $\mathcal{N}_{p_m}$ within the radius $r_m$ that are connected to the current center point $p_m = (x, y)$, as shown in (\ref{points_connectable}).
\begin{equation}
\label{points_connectable}
\mathcal{N}_{p_m} = \left\{ p_m \in \mathcal{V}_{GVD} \;\middle|\; \|p_m - p_m'\| < h,\;  \text{connected}(p_m, p_m') \right\}
\end{equation}
where $p_m$ denotes the current center point, and $p_m'$ represents the other points within the neighborhood $\mathcal{N}_{p_m}$ of radius $r_m$.

By integrating the map and the connectivity cache, only points that are unobstructed and reachable are included in the mean calculation and used to update the center point, as expressed in (\ref{mean}).
\begin{equation}
\label{mean}
p_m = \frac{1}{|\mathcal{N}_{p_m}|} \sum_{p_m‘ \in \mathcal{N}_{p_m}} p_m’
\end{equation}

If the new and old center points are unreachable, or if the new center point falls within an obstacle area, the iteration is terminated prematurely to ensure that the cluster center always remains in free space and is reachable. Convergent centers that are too close to each other are then merged to obtain a set of cluster centers $c$ that are reasonably distributed in space and non-overlapping. Finally, in the label assignment phase, the algorithm adopts the nearest neighbor principle, as defined in (\ref{label_rule}), to assign each sampling point to the nearest and connected cluster center. For each sample $\mathcal{V}_i$, the assigned label corresponds to the closest and connected cluster center.
\begin{equation}
\label{label_rule}
\ell_i = \arg\min_{k} \left\{ |\mathcal{V}_i - c_k| \mid \text{connected}(\mathcal{V}_i, c_k), c_k \in c \right\}
\end{equation}

As illustrated in the case 1 and case 2 of Fig. \ref{mean_shift}, compared to the traditional mean shift algorithm, the proposed method not only prevents cluster centers from being located on obstacles but also effectively suppresses the cross-structural node "attraction" phenomenon caused by obstacle barriers. This enhancement improves both the spatial coverage completeness and usability of the topological nodes. The overall process fully integrates spatial density with environmental reachability information, providing an accurate topological foundation for subsequent exploration tasks. Furthermore, the introduction of the KD-tree and LRU cache mitigates the computational burden associated with large-scale GVD nodes.

\subsubsection{Connection Of Topological Map Based On Switch Mechanism}

For the topological nodes obtained via mean-shift clustering, we propose a dual-layer switching connection strategy to construct a globally consistent and accurate environmental topological graph. First, both the clustered topological nodes and all GVD nodes are embedded into the topological graph. Based on the traversability of the GVD nodes, collision-free connectivity information between nodes is precomputed and stored using an LRU cache. Whenever multiple connected components appear in the topological graph, the smallest connected subgraph is identified, and the shortest GVD path between this subgraph and other nodes is queried. If such a path exists, an edge is directly added to achieve efficient GVD-based completion.
If no feasible GVD path is available, a fallback mechanism is triggered, employing a bidirectional A* algorithm to compute the shortest collision-free trajectory in the occupancy grid using nearest-neighbor heuristics. This trajectory is then converted into a topological edge. The resulting topological graph represents nodes as major passages or regions in the environment, with edges denoting feasible paths between these regions, forming an environmental topological representation based on spatial connectivity constraints.
By leveraging the switching mechanism between GVD and A*, the approach preserves the efficiency of GVD connectivity priors in large open spaces while utilizing the completeness of A* to ensure feasible connections in narrow or obstacle-dense areas. This achieves a balance among global topological integrity, real-time performance, and computational cost.

\begin{figure*}[t]
\centerline{\includegraphics[width=1\textwidth]{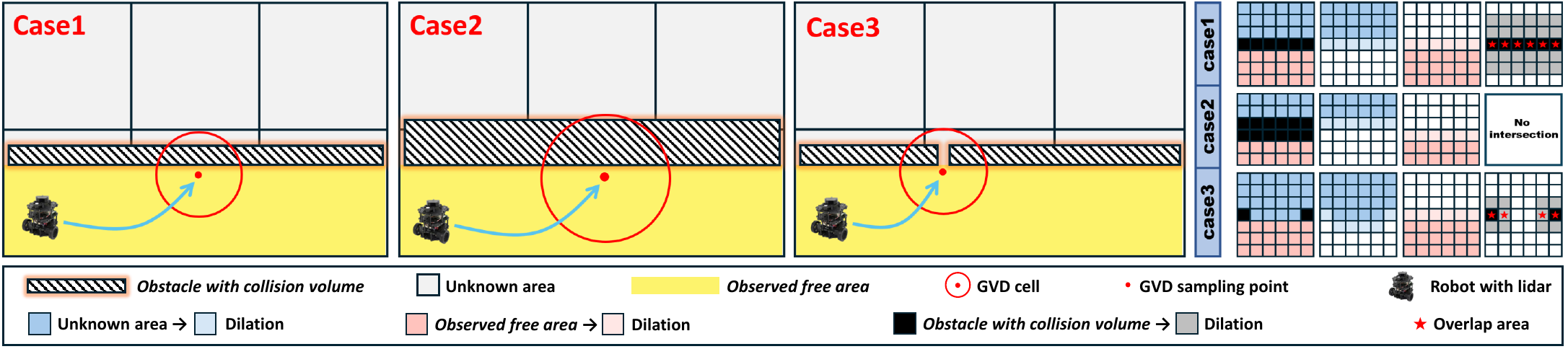}}
\caption{Frontier detection. The last figure shows the morphological dilation of unknown area, observed free area and obstacle respectively, and the calculation of the size of the overlapping area to determine whether it is frontier.}
\label{frontier_detection}
\end{figure*}

\begin{figure*}[t]
\centerline{\includegraphics[width=1\textwidth]{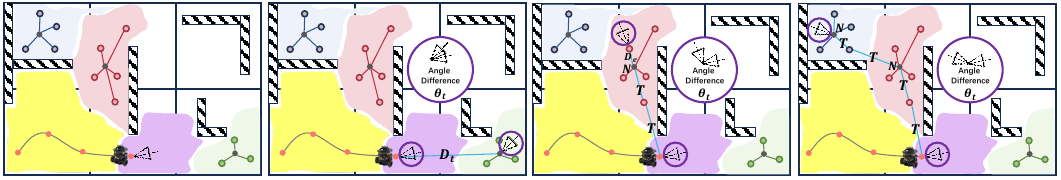}}
\caption{Frontier selection. The gray dots are topological nodes obtained by clustering GVD nodes, while the remaining dots are the original GVD nodes. When calculating cost, $D_t$ is the Euclidean distance between two nodes (when there is a straight path), while $T$, $N$ and $D_e$ respectively represent the total length of topological path to be detour, the number of topological nodes passed through and the distance to the end node. The last three subfigure respectively represent the cost for robot to evaluate the travel to three candidate target points in different color areas from the current position, and the one with the lowest cost will be selected as the next viewpoint.}
\label{cost_func}
\end{figure*}

\subsection{Viewpoints Selection Module}

\subsubsection{Frontier Detection Based on Morphology}


Unlike traditional frontier-based exploration methods, our method utilizes topological nodes and GVD nodes as navigation viewpoints, making the identification of frontier a critical challenge. Conventional frontier extraction methods determine frontier solely by detecting whether unknown grids exist in the GVD cell. As shown in Fig. \ref{frontier_detection}, when an unknown region is separated from the observed free area by obstacles and the node itself lies within the observed free region, designating such a node as the next navigation target will not yield new environmental information due to the obstruction, leading to repeated exploration. Therefore, additional verification is required to determine whether a valid passage for LiDAR beam exists between the unknown and known regions.

Specifically, if a GVD cell contains only unknown and free regions without any obstacles, the corresponding GVD node can be directly regarded as valid frontier. If obstacles are present, the reachability of the passage must be further evaluated. To this end, we propose the following frontier detection algorithm: first, a local submap is cropped from the global map using the center and radius of the candidate GVD node. Then, morphological dilation is applied separately to the unknown, free, and obstacle regions using an 8-neighborhood expansion to generate their respective neighborhood masks. Next, the number of overlapping pixels between the unknown and free regions is calculated via pixel-wise multiplication to extract the contact frontiers between the two regions. As shown in case 1 of Fig. \ref{frontier_detection}, if the number of overlapping pixels is minimal ($<5$), it indicates limited contact between the unknown and observed free regions, and the node is classified as a non-frontier point.

Furthermore, the obstacle neighborhood mask is multiplied with the overlapping region of the unknown and free areas, and the proportion of overlapping pixels relative to the total unknown–free overlap is calculated. As illustrated in case 2 and case 3 of Fig. \ref{frontier_detection}, when this proportion exceeds 0.8, it suggests that the unknown region is primarily adjacent to obstacle edges, representing an unreachable or pseudo-frontier with no practical exploration value. Such nodes should also be identified as explored non-frontier points. Only in cases outside the above conditions—where the unknown region sufficiently contacts the free region and is not concentrated near obstacles—is the node determined to be valid frontier.

This method effectively integrates spatial topology, morphological operations, and quantitative threshold criteria, successfully avoiding misjudgments caused by map noise or obstacle occlusion. It enhances the robustness and accuracy of frontier detection, providing a reliable basis for autonomous exploration tasks.

\subsubsection{Viewpoint Selection based on Designed Cost Function}


Upon completion of the topological graph construction and frontiers generation, the robot must select the optimal exploration target in real time from multiple candidate frontiers within the environment. To this end, we propose a lightweight cost evaluation framework that integrates topological, geometric, and kinematic information, balancing decision-making accuracy with computational efficiency.

Specifically, as shown in Fig. \ref{cost_func}, the framework first maps all candidate frontiers to their corresponding topological nodes in the constructed graph. Based on connectivity, it quickly determines whether an obstacle-free straight-line path exists between the robot’s current position and the target point, as expressed in (\ref{viewpoint_selection}). If such a path exists, the Euclidean distance $D_t$ is directly adopted as the base cost, with an additional weighting correction based on the heading angle $\theta_t$ to reduce motion losses caused by sharp turns. If no direct path exists, the topological path length $T$ is used as the main cost component, supplemented by a weighted combination of the terminal distance $D_e$, the number of topological nodes $N$, and the relative heading angle $\theta_t$, forming a comprehensive cost metric.

Here, the topological path length $T$ consists of two parts: the actual path length from the robot to the nearest topological node and the terminal distance $D_e$ from the topological endpoint to the target point. This design effectively captures the influence of obstacles on the actual path while avoiding excessive computational overhead from point-by-point path planning. The inclusion of the terminal distance $D_e$ compensates for spatial deviations between topological nodes and actual frontiers, ensuring accurate path estimation. The weighted node count $N$ encourages the robot to prioritize structurally simpler paths, thereby reducing the frequency of turns and replanning. The angle term $\theta_t$, leveraging the robot’s current orientation, further minimizes unnecessary rotational movements.

By integrating these factors, the proposed cost function maintains high accuracy in path length estimation while significantly reducing the time required for candidate point evaluation. This enhances the robot’s real-time exploration efficiency in complex and unknown environments.
\begin{equation}
\label{viewpoint_selection}
\begin{cases} 
C = D_t + \alpha \cdot \theta_t \cdot D_t, & \text{if } R \subseteq M_{free} \\ \\
C = T + \alpha \cdot N \cdot T + \beta \cdot \theta_t \cdot D_{e}, & \text{if } R \not \subseteq M_{free} \\
\end{cases}
\end{equation}

\section{Experiments}

\begin{figure}[t]
    \centering
    
    \begin{minipage}{0.24\linewidth}
        \centering
        \includegraphics[width=\textwidth]{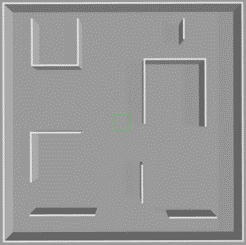}
        \subcaption{Env S1}
    \end{minipage}
    \hfill
    \begin{minipage}{0.24\linewidth}
        \centering
        \includegraphics[width=\textwidth]{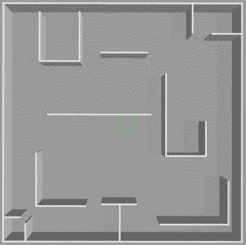}
        \subcaption{Env S2}
    \end{minipage}
    \hfill
    \begin{minipage}{0.24\linewidth}
        \centering
        \includegraphics[width=\textwidth]{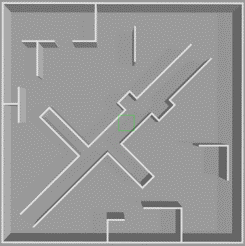}
        \subcaption{Env S3}
    \end{minipage}
    \hfill
    \begin{minipage}{0.24\linewidth}
        \centering
        \includegraphics[width=\textwidth]{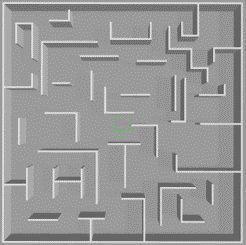}
        \subcaption{Env S4}
        \label{instrument}
    \end{minipage}
    
    \vspace{0.1cm}
    
    \begin{minipage}{0.32\linewidth}
        \centering
        \includegraphics[width=\textwidth]{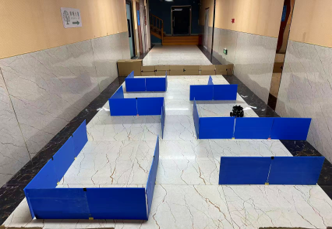}
        \subcaption{Env R1}
    \end{minipage}
    \hfill
    \begin{minipage}{0.32\linewidth}
        \centering
        \includegraphics[width=\textwidth]{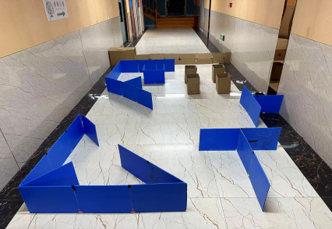}
        \subcaption{Env R2}
    \end{minipage}
    \hfill
    \begin{minipage}{0.32\linewidth}
        \centering
        \includegraphics[width=\textwidth]{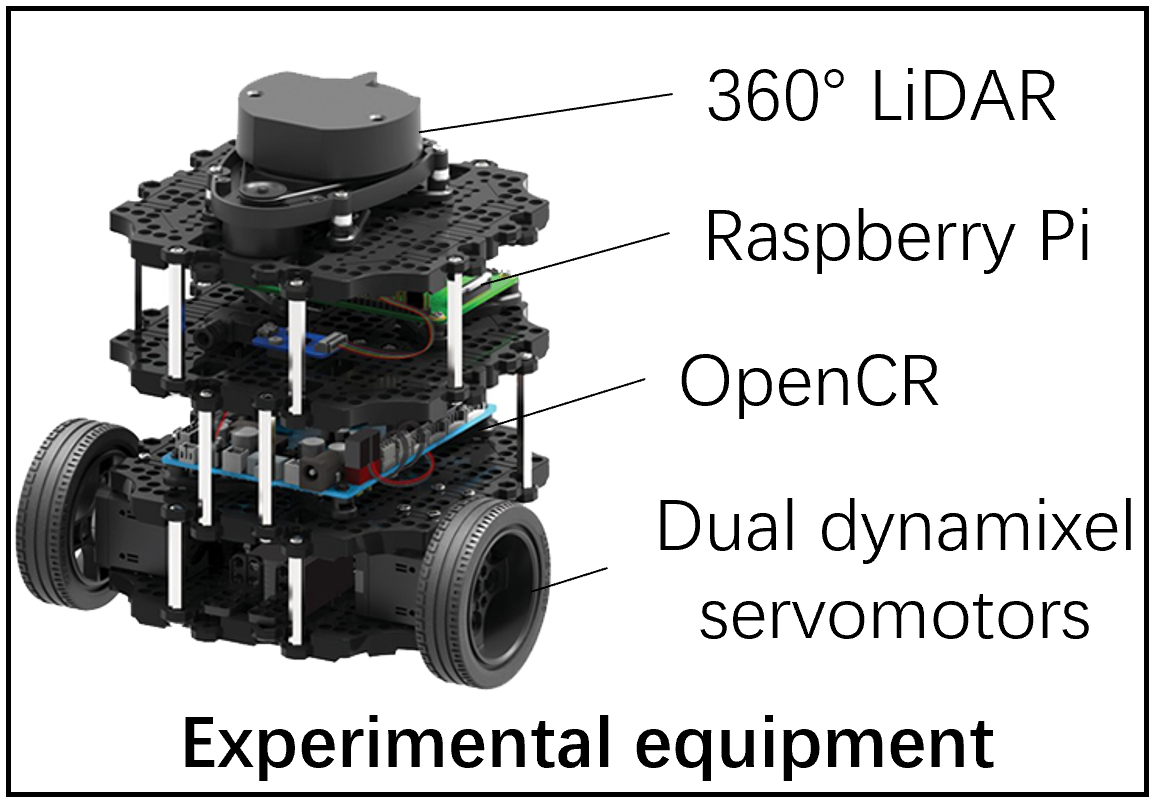}
        \subcaption{Robot}
        \label{robot}
    \end{minipage}
    
    \caption{Experimental environments and equipment.}
    \label{env_robot}
\end{figure}

We verify the exploration efficiency of the system respectively through simulations and real-world experiments. 
In simulations, we adopt a computer with 13th Gen Intel Core i7-13700K CPU and NVIDIA GeForce RTX2060 super GPU. The robot used in simulations is turtlebot3-burger.
In real-world experiments, we use a laptop with Intel i7-7700HQ CPU and NVIDIA GeForce GTX1060 GPU to remotely maneuver an turtlebot3-burger for validation. 
Turtlebot3-burger employs a Raspberry Pi 3 for high-level computation and an STM32F7-based OpenCR board for real-time motor control, complemented by dual dynamixel servomotors XL430-W250, a 360 laser distance sensor LDS-01, and an on-board IMU.
The environments of simulations are shown as S1-S4 in Fig. \ref{env_robot} and the environments of real-world experiments are shown as R1-R2 in Fig. \ref{env_robot}. In addition, experimental setup is shown in Table. \ref{setup}, and experimental equipment is shown in Fig. \ref{robot}.

\begin{table}[t]
    \caption{Experimental setup.}
   
    \label{setup}
    \centering
    \resizebox{1\columnwidth}{!}{
    \begin{tabular}{c|c| cccccc}
      \toprule[0.5mm]
      \textbf{Type} & \textbf{Scene} & \textbf{Max Speed(m/s)} & \textbf{Max Angle(°/s)}  & \textbf{Lidar range(°)} & \textbf{Termination condition}\\
      
      \midrule
      \midrule
      
      \multirow{4}{*}{\rotatebox{90}{\textit{\textbf{Simulation}}}}
      &S1 & 0.3 & 17.1  
      &  270 & 95\%  \\
      
      &S2 & 0.3 &  17.1 
      &  270 & 95\%  \\
      
      &S3 & 0.3 &  17.1 
      &  270 & 95\%  \\

      &S4 & 0.3 & 17.1 
      &  270 & 95\%  \\

      \midrule
      \midrule
      \multirow{2}{*}{\rotatebox{90}{\textit{\textbf{Real-}}}}
      \multirow{2}{*}{\rotatebox{90}{\textit{\textbf{World}}}}
      &R1 & 0.3 & 18.9  &  360 & 95\%   \\
      &R2 & 0.3 & 18.9  &  360 & 95\%   \\
      
      \bottomrule[0.5mm]
      
    \end{tabular}}
   
\end{table}





\subsection{Simulations}

\begin{figure}[t]
\centerline{\includegraphics[width=0.5\textwidth]{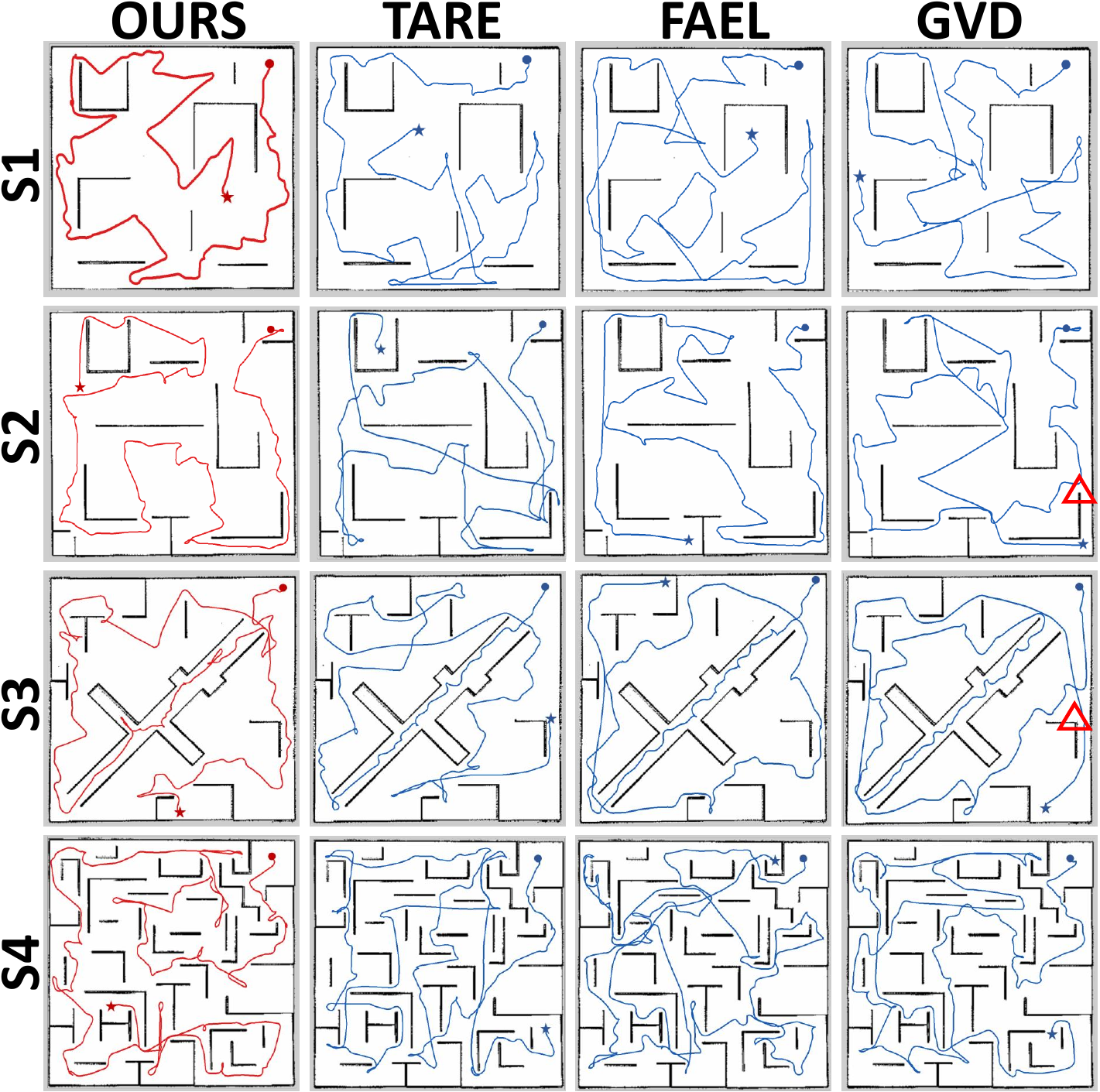}}
\caption{Trajectory of simulations. The dots represents starting point, and star represents ending point.}
\label{simulation_traj}
\end{figure}

To evaluate the exploration efficiency of the system, simulation experiments are conducted in four different simulation environments. As shown in Table. \ref{simulation_ex}, by comparing two metrics (exploration time and movement distance) with three methods (TARE\cite{cao2021tare}, FAEL\cite{10015689}, GVD\cite{10771660}), each method is executed ten times at the same location in each environment. In Fig. \ref{simulation_traj}, we also present the trajectory routes of each method to compare the path backtracking frequencies.

In S1 (20m $\times$ 20m), we deploy a small number of obstacles. Given the simplicity of this scenario, all methods achieve 100\% success rate. Compared to other methods, our method reduces exploration time by up to 36.9\%. Moreover, as shown in the Fig. \ref{simulation_traj}, the trajectory of our method exhibits no overlapping points and avoids path backtracking. 
In S2 (20m $\times$ 20m), we further complicate the environmental structure. As illustrated in the Fig. \ref{simulation_traj}, GVD\cite{10771660} fails to enter the narrow passage at the red marked location and returns to the same point during the final phase of the task, resulting in path backtracking. In contrast, our method accurately captures narrow passages through hierarchical GVD sampling, thereby maximizing exploration efficiency. Compared to other methods, our method reduces exploration time by up to 23.9\%. 
In S3 (20m $\times$ 20m), we add a bidirectional passage to partition the area. As seen in the Fig. \ref{simulation_traj}, GVD\cite{10771660} again misses the narrow passage at the red marked location, leading to repeated path backtracking. Our method reduces exploration time by up to 27.9\% compared to GVD\cite{10771660}. 
In S4 (20m $\times$ 20m), we configure an extremely complex obstacle structure, making the environment highly challenging for exploration tasks. In this environment, our method achieves 100\% success rate, while other methods only achieve success rates between 40\% - 60\%. The trajectory of our method generally follows a trend of circumventing the environment, unlike other methods that frequently traverse the environment arbitrarily, resulting in frequent path backtracking. Compared to other methods, our method reduces exploration time by up to 25.8\%.

Our method demonstrates significant advantages in exploration efficiency and path planning quality across various environments. Furthermore, our method maintains perfect success rates in challenging scenarios where other methods deteriorate to 40-60\% success rates, proving its superior robustness in complex obstacle configurations.

\begin{table}[t]
\centering
\caption{Result of simulations.}
\label{simulation_ex}
\resizebox{\columnwidth}{!}{%
\begin{tabular}{c|c|cc|cc|c|c|c}
\toprule[0.5mm]
\multirow{2}{*}{Scene} &
  \multirow{2}{*}{Method} &
  \multicolumn{2}{c|}{\begin{tabular}[c]{@{}c@{}}Exploration\\ Time(s)\end{tabular}} &
  \multicolumn{2}{c|}{\begin{tabular}[c]{@{}c@{}}Movement\\ Distance(m)\end{tabular}} &
  \multirow{2}{*}{\begin{tabular}[c]{@{}c@{}}Average\\ Speed(m/s)\end{tabular}} &
  \multirow{2}{*}{\begin{tabular}[c]{@{}c@{}}Time-limited\\ Number(s)\end{tabular}} &
  \multirow{2}{*}{\begin{tabular}[c]{@{}c@{}}Success \\ Number\end{tabular}} \\ \cline{3-6}
  
&     & \multicolumn{1}{c|}{avg} & std & \multicolumn{1}{c|}{avg} & std &  &  &  \\ \midrule[0.5mm]
                    
\multirow{4}{*}{S1} 
&Ours &\multicolumn{1}{c|}{287.3} & 10.7  &\multicolumn{1}{c|}{86.19}  & 16.9  & 0.3 & 500 & 10 \\  
&TARE\cite{cao2021tare} &\multicolumn{1}{c|}{450.6} & 21.6 &\multicolumn{1}{c|}{112.65} & 18.1 & 0.25 & 500 & 10 \\ 
&FAEL\cite{10015689} &\multicolumn{1}{c|}{455.1} & 18.4 &\multicolumn{1}{c|}{127.4}  & 19.2 & 0.28 & 500 & 10 \\ 
&GVD\cite{10771660} &\multicolumn{1}{c|}{414.1} & 25.2 &\multicolumn{1}{c|}{111.8} & 20.5 & 0.27 & 500 & 10 \\ \midrule[0.5mm]

\multirow{4}{*}{S2} 
&Ours &\multicolumn{1}{c|}{372.3} & 14.4 &\multicolumn{1}{c|}{111.6} & 16.8 & 0.3 & 550 & 10 \\ 
&TARE\cite{cao2021tare} &\multicolumn{1}{c|}{464.5} & 23.7 &\multicolumn{1}{c|}{130.1} & 22.8 & 0.28 & 550 & 10 \\ \
&FAEL\cite{10015689} &\multicolumn{1}{c|}{446.5} & 17.2 &\multicolumn{1}{c|}{120.6} & 18.5 & 0.27 & 550 & 10 \\ 
&GVD\cite{10771660} &\multicolumn{1}{c|}{489.5} & 35.1 &\multicolumn{1}{c|}{122.4} & 25.3 & 0.25 & 550 & 10 \\ \midrule[0.5mm]

\multirow{4}{*}{S3} 
&Ours &\multicolumn{1}{c|}{401.3} & 16.5 &\multicolumn{1}{c|}{120.4} & 18.2 & 0.3 & 600 & 10 \\ 
&TARE\cite{cao2021tare} &\multicolumn{1}{c|}{416.3}  & 33.1 &\multicolumn{1}{c|}{108.2} & 22.1 & 0.26 & 600 & 10 \\  
&FAEL\cite{10015689} &\multicolumn{1}{c|}{454.3} & 27.2 &\multicolumn{1}{c|}{122.6} & 21.8 & 0.27 & 600 & 10 \\ 
&GVD\cite{10771660} &\multicolumn{1}{c|}{556.3} & 29.6 &\multicolumn{1}{c|}{144.6} & 25.7 & 0.26 & 600 & 8 \\ \midrule[0.5mm]

\multirow{4}{*}{S4} 
&Ours &\multicolumn{1}{c|}{523.3} & 20.2 &\multicolumn{1}{c|}{151.8} & 21.2 & 0.29 & 750 & 10 \\ 
&TARE\cite{cao2021tare} &\multicolumn{1}{c|}{639.3} & 35.8 &\multicolumn{1}{c|}{159.9} & 28.9 & 0.25 & 750 & 4 \\ 
&FAEL\cite{10015689} &\multicolumn{1}{c|}{705.2} & 32.4 &\multicolumn{1}{c|}{183.4} & 27.8 & 0.26 & 750 & 6 \\  
&GVD\cite{10771660} &\multicolumn{1}{c|}{665} & 41.7 &\multicolumn{1}{c|}{172.9} & 33.5 & 0.26 & 750 & 4 \\ \bottomrule[0.5mm]
\end{tabular}%
}
\end{table}

\subsection{Real-World Experiments}

\begin{figure}[t]
\centerline{\includegraphics[width=0.5\textwidth]{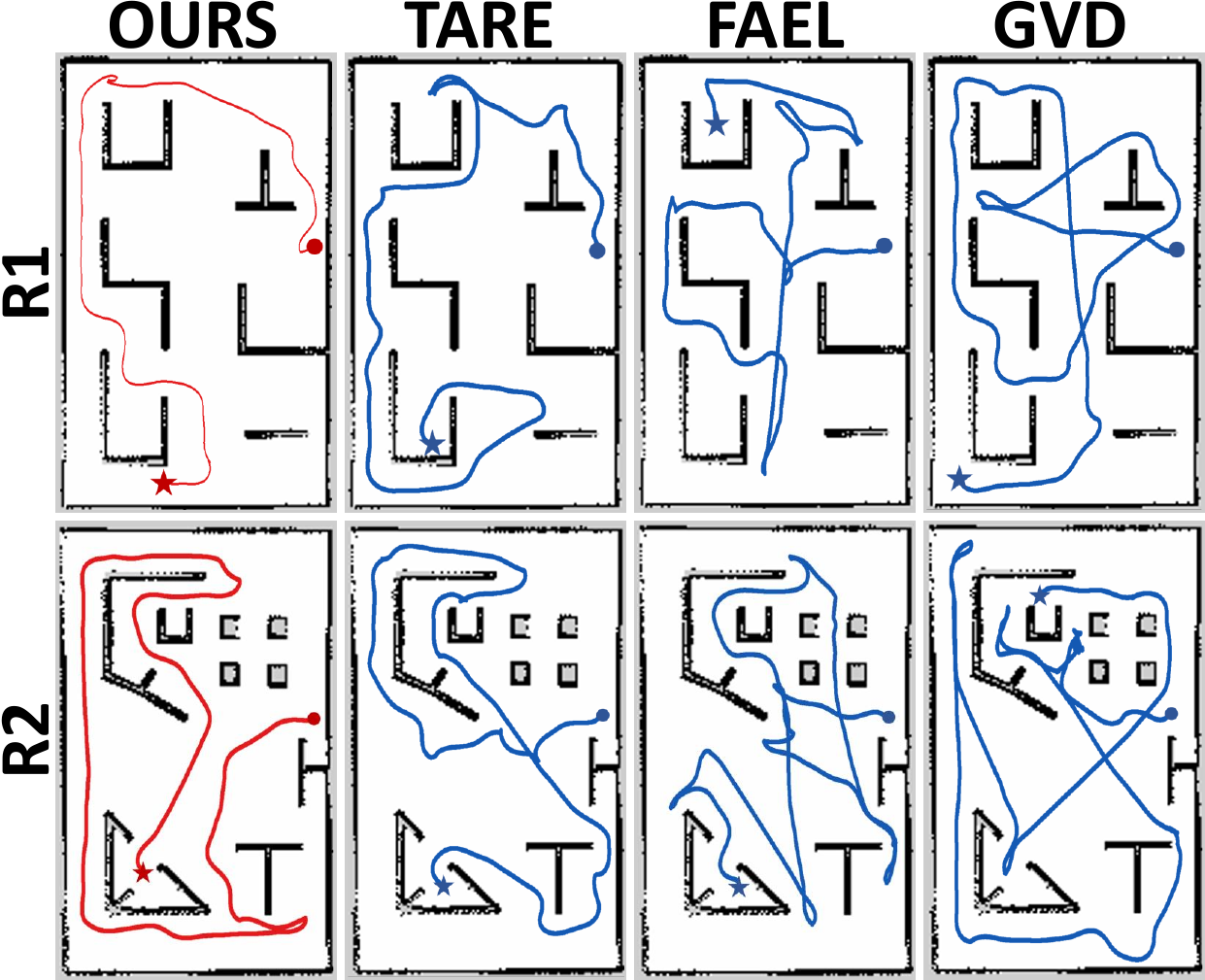}}
\caption{Trajectory of real-world experiments. The dots represents starting point, and star represents ending point.}
\label{real_traj}
\end{figure}

We also conduct real-world experiments in two different real-world environments. As shown in Table. \ref{real_ex}, by comparing two metrics (exploration time and movement distance) with three methods (TARE\cite{cao2021tare}, FAEL\cite{10015689}, GVD\cite{10771660}), each method is executed five times at the same location in each environment. In Fig. \ref{real_traj}, we also present the trajectory routes of each method to compare the path backtracking frequencies.

In R1 (7m $\times$ 3m), all methods achieved 100\% success rate. Our method consumed the shortest time. Compared with other methods, the exploration time of ours is reduced by up to 71.7\%. As can be seen from Fig. \ref{real_traj}, both FAEL\cite{10015689} and GVD\cite{10771660} experience varying degrees of path backtracking. This leads to a relatively low exploration efficiency in actual scenarios. 
In R2 (7m $\times$ 3m), the success rate of GVD\cite{10771660} is only 60\%, while our method could still achieve 100\% exploration success rate. Compared with other method, the exploration time of ours is reduced by up to 46.4\%.

In actual scenario tests, our method has demonstrated significant advantages, with the exploration efficiency increasing by up to 71.7\% and maintaining 100\% task success rate consistently. It has shown stronger practicality and reliability of our method in real and complex environments.

\begin{table}[t]
\centering
\caption{Result of real-world experiments.}
\label{real_ex}
\resizebox{\columnwidth}{!}{%
\begin{tabular}{c|c|cc|cc|c|c|c}
\toprule[0.5mm]
\multirow{2}{*}{Scene} &
  \multirow{2}{*}{Method} &
  \multicolumn{2}{c|}{\begin{tabular}[c]{@{}c@{}}Exploration\\ Time(s)\end{tabular}} &
  \multicolumn{2}{c|}{\begin{tabular}[c]{@{}c@{}}Movement\\ Distance(m)\end{tabular}} &
  \multirow{2}{*}{\begin{tabular}[c]{@{}c@{}}Average\\ Speed(m/s)\end{tabular}} &
  \multirow{2}{*}{\begin{tabular}[c]{@{}c@{}}Time-limited\\ Number(s)\end{tabular}} &
  \multirow{2}{*}{\begin{tabular}[c]{@{}c@{}}Success \\ Number\end{tabular}} \\ \cline{3-6}
  
&     & \multicolumn{1}{c|}{avg} & std & \multicolumn{1}{c|}{avg} & std &  &  &  \\ \midrule[0.5mm]
                    
\multirow{4}{*}{R1} 
&Ours &\multicolumn{1}{c|}{35.1} & 10.3  &\multicolumn{1}{c|}{10.5}  & 3.7  & 0.3 & 150 & 5 \\  
&TARE\cite{cao2021tare} &\multicolumn{1}{c|}{49.1} & 16.1 &\multicolumn{1}{c|}{14.2} & 5.9 & 0.29 & 150 & 5 \\ 
&FAEL\cite{10015689} &\multicolumn{1}{c|}{52.6} & 12.3 &\multicolumn{1}{c|}{15.8}  & 5.2 & 0.3 & 150 & 5 \\ 
&GVD\cite{10771660} &\multicolumn{1}{c|}{123.9} & 14.5 &\multicolumn{1}{c|}{18.6} & 3.2 & 0.15 & 150 & 5 \\ \midrule[0.5mm]

\multirow{4}{*}{R2} 
&Ours &\multicolumn{1}{c|}{51.3} & 12.4 &\multicolumn{1}{c|}{15.4} & 5.1 & 0.3 & 150 & 5 \\ 
&TARE\cite{cao2021tare} &\multicolumn{1}{c|}{67.3} & 23.7 &\multicolumn{1}{c|}{15.5} & 5.2 & 0.23 & 150 & 4 \\ 
&FAEL\cite{10015689} &\multicolumn{1}{c|}{70.5} & 20.5 &\multicolumn{1}{c|}{19.7} & 6.9 & 0.28 & 150 & 5 \\  
&GVD\cite{10771660} &\multicolumn{1}{c|}{95.7} & 15.4 &\multicolumn{1}{c|}{25.9} & 7.8 & 0.27 & 150 & 3 \\ \bottomrule[0.5mm]
\end{tabular}%
}
\end{table}

\section{conclusion}

This paper proposed a topological map construction method based on hierarchical GVD sampling, which effectively solved the key problems such as frequent path backtracking, missed detection of narrow channels, and unreachable topological nodes when robots conducted autonomous exploration in unknown and complex environments. 
Firstly, a hierarchical GVD sampling strategy with coverage awareness was proposed. By conducting global-local dual-granularity sampling and coverage map maintenance for the denoised map, efficient capture of environmental features and effective suppression of redundant nodes were achieved. 
Secondly, a connection-constrained mean shift clustering algorithm was designed, combined with KD-tree acceleration and a double-layer switching connection mechanism, to ensure the reachability of topological nodes and the connectivity of the map. 
Finally, a frontier detection method based on morphological dilation was introduced, combined with a lightweight cost function, to achieve precise screening of real exploration targets and selection of the optimal viewpoint. 
The simulation and real-world experiment results showed that this method can maintain 100\% exploration success rate in various complex scenarios, with the exploration time reduced by up to 71.7\%, and there was very little backtracking phenomenon. It was significantly superior to the comparison methods, fully verifying the superiority of our method in improving exploration efficiency and robustness.



\bibliographystyle{Bibliography/IEEEtranTIE}
\bibliography{Bibliography/BIB_xx-TIE-xxxx}\ 

\end{document}